\documentclass[conference]{IEEEtran}
\pdfoutput=1

\usepackage{cite}
\usepackage{amsmath,amssymb,amsfonts}
\usepackage{graphicx}
\usepackage{textcomp}
\usepackage{xcolor}

\usepackage{subcaption}
\usepackage{caption}

\usepackage{threeparttable}

\usepackage{algorithm}
\usepackage[noend]{algpseudocode}
\usepackage{mathtools}

\def\BibTeX{{\rm B\kern-.05em{\sc i\kern-.025em b}\kern-.08em
    T\kern-.1667em\lower.7ex\hbox{E}\kern-.125emX}}

\begin{document}


\title{Experience Sharing Between Cooperative Reinforcement Learning Agents \thanks{Published at the Proceedings of the 31$^{st}$ IEEE International Conference on Tools with Artificial Intelligence.}
}


\author{\IEEEauthorblockN{Lucas Oliveira Souza}
\IEEEauthorblockA{
\textit{Numenta}\\
Redwood City, USA \\
lsouza@numenta.com}
\and
\IEEEauthorblockN{Gabriel de Oliveira Ramos}
\IEEEauthorblockA{\textit{Graduate Program in Applied Computing} \\
\textit{Universidade do Vale do Rio dos Sinos}\\
S\~ao Leopoldo, Brazil \\
gdoramos@unisinos.br}
\and
\IEEEauthorblockN{Celia Ghedini Ralha}
\IEEEauthorblockA{\textit{Computer Science Department} \\
\textit{University of Brasilia}\\
Brasilia, Brazil \\
ghedini@unb.br}
}

\maketitle

\begin{abstract}
The idea of experience sharing between cooperative agents naturally emerges from our understanding of how humans learn. Our evolution as a species is tightly linked to the ability to exchange learned knowledge with one another. It follows that experience sharing (ES) between autonomous and independent agents could become the key to accelerate learning in cooperative multiagent settings. We investigate if randomly selecting experiences to share can increase the performance of deep reinforcement learning agents, and propose three new methods for selecting experiences to accelerate the learning process. Firstly, we introduce Focused ES, which prioritizes unexplored regions of the state space. Secondly, we present Prioritized ES, in which temporal-difference error is used as a measure of priority. Finally, we devise Focused Prioritized ES, which combines both previous approaches. The methods are empirically validated in a control problem. While sharing randomly selected experiences between two Deep Q-Network agents shows no improvement over a single agent baseline, we show that the proposed ES methods can successfully outperform the baseline. In particular, the Focused ES accelerates learning by a factor of 2, reducing by 51\% the number of episodes required to complete the task.
\end{abstract}



\section{Introduction}
\label{sec:introduction}


Learning from experience sharing is a core component of human society. Humans do not need to learn everything from scratch. While learning from experience, we also exchange knowledge with peers and teachers to accelerate the learning process. Thus, acquiring knowledge involves as much information transfer as it involves discovery by trial-and-error.

This intuition can be extended to multiagent scenarios with cooperative agents, where agents are either attempting to achieve a common goal or coexisting in the environment while pursuing individual goals. If cooperation is done intelligently, each agent can benefit from other agents' instantaneous information, episodic experience, or learned knowledge \cite{tan1993multi}. 

Sharing experiences in Reinforcement Learning (RL) agents was first investigated in \cite{tan1993multi, whitehead1991complexity}. One of the main issues in learning by trial-and-error is that it relies on the agent's luck in first achieving the goal by chance, which could be overcome by learning a policy directly from external experts \cite{lin1992self}. 
In this context, Tan \cite{tan1993multi} proposed two knowledge sharing approaches: 
sharing a learned policy, between a more knowledgeable agent and a novice one; and sharing experiences, tuples that represent the state, action, reward received, and the next state which the agent is transitioned to.

Sharing a part of the policy, in the form of action advice, is explored in the teacher-student framework introduced in  \cite{torrey2013teaching}, and extended by \cite{da2017simultaneously}, which considers the possibility of all agents acting both as teachers and learners in a multiagent setting. When faced by a situation in which it has low confidence in its policy, agents may request other cooperative agents for help. Action advice has the advantage of being easily extendable to heterogeneous agents, but requires instant communication between the agents and limits the amount of knowledge exchanged per communication to the current transition. 

In this work, nonetheless, we follow the second approach proposed in \cite{tan1993multi}, namely experience sharing (ES). Experiences can be shared by batch and at sparse intervals, reducing the communication overhead between agents. 
Sharing experiences between heterogeneous agents can be done by priorly evaluating if the agents are similar enough to benefit from the knowledge to be shared \cite{verstraetenreinforcement}. A similar process can be used in heterogeneous environments, but using a distance function to determine the similarity of the environments the agents are currently in \cite{garant2017context}.

In our proposal, we investigate ES among agents concurrently learning a similar task. Each agent learns a policy independently from other agents, and interacts with them only for the purpose of sharing experiences. We first investigate the premise that sharing experiences alone is enough to increase learning performance of the agents. We then propose a method which limits shared experiences to those that are novel to the learning agent. While naive ES between two agents shows no improvement over single agent learning, our proposed Focused ES method is able to achieve a 51.4\% reduction in the number of episodes required to complete a task. 


The remaining of this paper is organized as follows. In Section 2, we provide a background on RL and recent advances in Deep Reinforcement Learning (DRL) related to the proposal. In Section 3, we present our proposal. In Section 4, we detail the experiments conducted and discuss results. In Section 5, we do a brief review of related work. Finally, in Section 6, we present our conclusions and discuss future research directions. 

\section{Reinforcement Learning}

In this section, we discuss the reinforcement learning problem, the Q-learning algorithm and its combination with neural networks, and some of the latest improvements.

An agent perceives the world through sensors and changes it through its actions. Each action taken by the agent affects the environment, that may output a reward. RL involves learning what to do, mapping from situations to actions in a given environment in order to maximize a numerical reward signal \cite{sutton1998reinforcement}.

The RL problem is commonly modeled as a Markov Decision Process (MDP), formalized by a set of states $S$, a set of actions $A$, a transition probability function 
\begin{equation}
    p(s' \mid s,a) = Pr \{ s_{t+1} = s' \mid s_t = s, a_t = a \},
\end{equation} 
and a reward function 
\begin{equation}
    r(s,a) = \mathbb{E} \{R_{t+1} \mid s_t = s, a_t = a\}.
\end{equation} 
The agent moves from one state to another through its actions. A transition probability function determines which next state $s'$ the agent arrives after taking action $a$. After arriving at the new state, the agent receives a reward, which can be null, positive or negative \cite{sutton1998reinforcement}.

The goal of RL in this MDP setting is to learn an optimal policy  $\pi^{*}(s) \rightarrow a$, which maximizes $\sum_{t \geq 0} \gamma^t r_t$. Policy is a function that determines which action the agents needs to take given the perceived state. If we consider a finite amount of time $T$, every sequence of actions from the agent from time 0 to time $T$ is considered an episode. An agent thrives to maximize not only local reward but the total reward for an episode. The total reward can either be spread upon intermediate states or concentrated in the final state, introducing the problem of learning an optimal policy in a delayed rewards setting. 

RL problems can involve one or more agents. Multiagent settings can be divided between fully cooperative, fully competitive or somewhere in between, which comprises a wide spectrum of scenarios. We are interested in problems with cooperative agents, where agents learn concurrently to achieve a similar but independent goal. Multiagent systems can benefit from the speed of parallel computation, experience sharing by communication, teaching or imitation \cite{busoniu2008comprehensive}.

\subsection{Q-Learning and Deep Q-Network}

Q-Learning is a RL model-free algorithm where an agent attempts to learn an action-value function $Q^{*}(s,a)$ \cite{watkins1992q}. The agent experiences the world by choosing an action $a$ at each state $s$, reaching the next state $s_{t+1}$ and perceiving a reward $r$. The action-value is updated based on the perceived reward plus the expected reward from the future states, discounted by $\gamma$. The update rule of the Q-function is given by:

\begin{equation}\label{eq-qlearning}
    \delta = {r_{t}} + \gamma \max_{a} Q(s_{t+1},a) - Q(s_{t},a_{t})
\end{equation}
\begin{equation}
    Q(s_{t},a_{t}) = Q(s_{t},a_{t}) + \alpha \delta.
\end{equation}

In continuous state space environments, it is infeasible to represent the Q-value function as a table, requiring it to be approximated. The Deep Q-Network (DQN) algorithm \cite{mnih2015human} was able to successfully use neural networks as function approximators to the Q-value functions, using experience replay and a separate target network to stabilize learning and avoid overfitting, issues faced in past attempts. The introduction of DQN ignited the field of DRL, leading to outstanding results and the development of a wider class of algorithms using neural networks as function approximators. 

\subsection{Experience Replay}

Experience replay (ER) was first introduced in \cite{lin1991programming}, in which the author notes some experiences may be rare and costly to acquire and points to the inefficiency of discarding experiences obtained through RL after they are used only once. The experiments conducted by Lin are prescient and in many ways very close to the DQN algorithm that jump-started the field of DRL \cite{lin1991programming, lin1992self, mnih2015human}. 

ER has also been described as an effective approximator to model-based RL algorithms \cite{lin1992self, vanseijen2015deeper, altahhan2018td}. Instead of learning a parameterized model to generate transitions, the agent samples from past transitions, leading to similar results. ER was sporadically addressed in the literature in the following decade, mainly as an additional methodology for data efficiency in complex domains \cite{smart2000practical, kalyanakrishnan2007batch, adam2012experience}. But it has resurfaced with DQN and can be considered the most important gradient of the modern DRL algorithms that achieved major breakthroughs in recent years \cite{mnih2015human, lillicrap2015continuous}. 

In prior work, experiences were either sampled backward \cite{lin1991programming}, or more commonly sampled randomly. Random sampling maintains the premise of independent identically distributed samples, required to guarantee convergence of the gradient descent algorithm \cite{mnih2015human}. Empirical experimentation, however, showed that attributing an importance score to each experience and using it to steer the sampling process to focus on specific experiences lead to greater data efficiency in DRL algorithms \cite{schaul2015prioritized}.

Improving the diversity of experience buffer has also been shown to increase performance in DRL algorithms \cite{ong2015distributed,nair2015massively}, where multiple agents are trained in parallel, but share a common set of value function parameters and a common replay buffer. In \cite{horgan2018distributed} is shown that having a single ER buffer shared amongst agents, without sharing the value function, is alone sufficient to improve the results of DQN.

The benefits of a more diverse experience buffer can be extended to cooperative multiagent scenarios, where agents are independent and autonomous, by allowing agents to share experiences with one another during the learning process. This motivation is the core principle behind our Focused ES proposal.

\section{Proposal}


In all our scenarios, two or more agents learn concurrently. The agents have independent MDPs with similar goals. Experience is defined as a tuple $(s_t, a_t, r_t, s_{t+1})$, representing a transition taken by an agent from one state to another and the response received from the environment. All methods can be applied to any model-free DRL algorithm which makes use of experience replay, including the popular DQN \cite{mnih2015human} and Deep Deterministic Policy Gradient (DDPG) \cite{lillicrap2015continuous}.




We divide the ES process into two stages. In the episode stage, each agent incorporates experiences received from the last sharing stage into its buffer and executes the episode. After completing the episode, the agent issues a new request for help to a public requests board. The stage ends when all agents have completed their episodes. In the sharing stage, all agents alternatively assume the role of teacher. As a teacher, the agent fulfills others' requests on the request board by 
sending a batch of experiences to the requesting agent's inbox, with the batch size limited to the minimum between $\kappa$ and the experience buffer size, where $\kappa$ is a hyperparameter of the model.

\begin{figure*}[!t]
\centering
  \includegraphics[width=16.1cm]{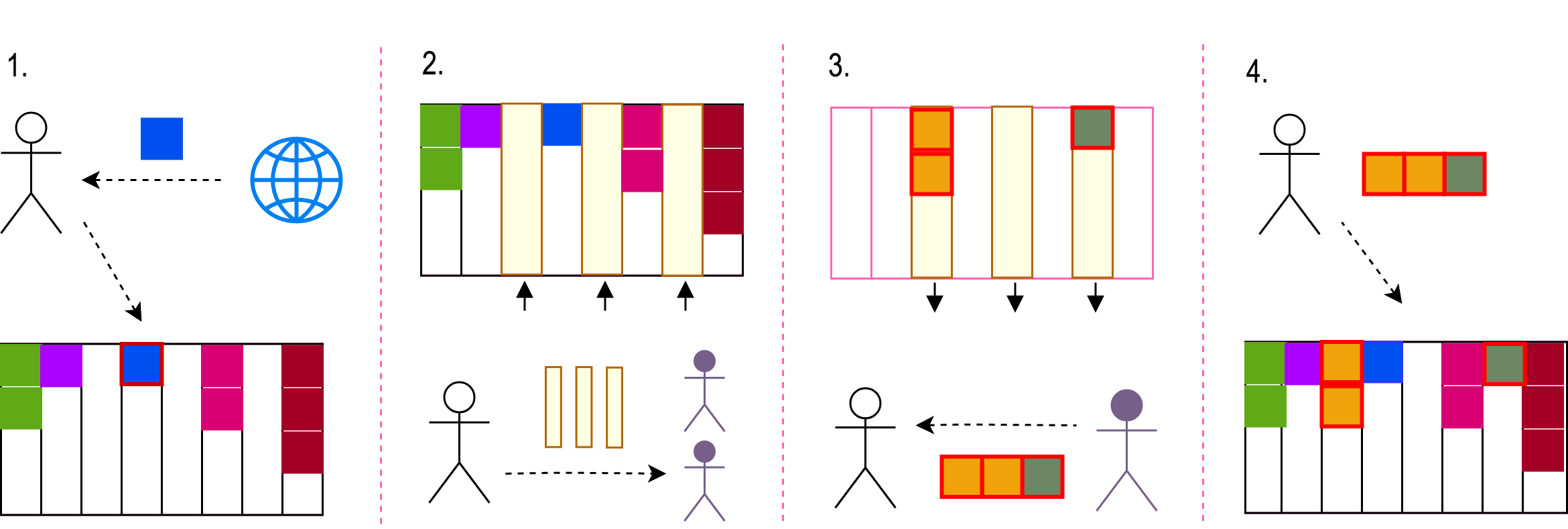}\\
  \caption{Schematics of the Focused Experience Sharing method.}
\label{fig:schematics}
\end{figure*}


ES is typically performed by \textit{frequently} sharing \textit{small} batches of experiences. By contrast, we will focus on the episode by episode approach, allowing for a more varied range of experiences to be included in the batch before the sharing occurs and limiting the communication between the agents to once each episode. The pseudocode is described in Alg. \ref{algorithm:sharing}.

\begin{algorithm}[H]
\caption{Experience sharing} 
\label{algorithm:sharing}
\begin{algorithmic}[1]
\State initialize environment and agents
\State initialize empty requests board \textit{RB}
\While{\textbf{not} (all agents completed)}
    \For{agent $\mathcal{A}$ in the environment}
        \State $\mathcal{A}$ add experiences from inbox to buffer
        \State $\mathcal{A}$ plays episode
        \State $\mathcal{A}$ adds new request $\mathcal{R}$ to \textit{RB} 
    \EndFor
    \For{agent $\mathcal{A}$ in the environment}
        \State check \textit{RB} for available requests from other agents
        \For{request $\mathcal{R}$ in available requests}
            \State $\mathcal{A}$ sample batch of experiences $\mathcal{B}$ matching $\mathcal{R}$
            \State $\mathcal{A}$ places $\mathcal{B}$ in requesting agent's inbox
        \EndFor
    \EndFor
    \State clear \textit{RB}
\EndWhile
\end{algorithmic}
\end{algorithm}
We propose four methods of ES, and proceed to validate them empirically. The methods differ mainly in how the request is composed by the requesting agent, and how experiences to be shared are selected by the teacher agent. In the sequence the methods are detailed. 

\subsection{Naive ES} 

Requests contain no details. Experiences shared are randomly sampled from the teacher's buffer.

\subsection{Prioritized ES} 

Requests contain no details. Experiences to share are sampled using priorities to define the probability of an experience being sampled. The priorities used for ES are the same defined in the Prioritized Replay method \cite{schaul2015prioritized}. The priority of an experience is defined as the temporal difference error (TD-error) calculated when the experience is used for learning. The TD-error is the difference between the total return expected to be obtained from the experience and the actual return obtained. It can also be understood as a measure of surprise, or how unexpected the experience is to the agent that lives it. Since the teacher agent has no access to the student's other than the request, it calculates priorities based on its own action-value function.

\subsection{Focused ES}  

Request contains details regarding which regions of the state space are poorly explored. When forming the request, the agent swipes its buffer to identify regions of the state space which contains fewer experiences. This can be achieved by maintaining a second structure parallel to the buffer, an occupancy grid. Whenever a new experience is added to the buffer, the agent discretizes the state using state aggregation, which consists of binning each continuous variable and combining the resulting bins. The experience is allocated to the occupancy grid according to the discretized state and the action. 

Storing an experience corresponds to step one in the schematic process shown in Fig. \ref{fig:schematics}. In step two, the agent selects a mask of the occupancy grid where each position is marked as unexplored if the number of experiences in the position is less or equal a threshold $\zeta$. By varying $\zeta$ we can control for how many experiences define what it means for a region to be unexplored. 

In step three, the teacher agents who receive the request use the request mask to identify experiences in its buffer that belongs to the unexplored regions of the student's state space. This procedure requires both agents to have exactly similar state and action space, being suitable only to homogeneous agents in similar environments. The experiences selected in step three are randomly sampled to form the batch of experiences to be sent to the inbox of the requesting agent, which are later added to its buffer (step four).

\subsection{Prioritized Focused ES}  

Combines the Focused ES and Prioritized ES methods. The only change to the Focused ES method is in the last step. Instead of randomly sampling from the experiences selected, select them based on the priorities defined in Prioritized ES.



\section{Experiments}

The ES methods are evaluated empirically in simulated environments. We describe the baseline algorithm and environment before proceeding for the results and discussion.

\subsection{Environment}

We evaluated the experiment in the Cart Pole environment, a classic control problem introduced in \cite{sutton1998reinforcement}, using the OpenAI Gym library \cite{brockman2016openai}. 

\begin{figure}[t]
\centering
  \includegraphics[width=.7\linewidth]{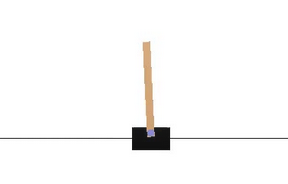}\\
  \caption{OpenAI CartPole environment}
  \label{fig:cartpole}
\end{figure}

Cart Pole environment, represented in Fig. \ref{fig:cartpole}, consists of balancing a pole, attached by an un-actuated joint to a cart, that moves along a frictionless track. The goal of the agent is to apply force to the cart, so as to balance a pendulum standing on top of it. There are two discrete actions available, which corresponds to either applying a force of +1 to move the cart to the right or a force of -1 to move the cart to the left (there is also a version of this environment with continuous action space, which we will not cover in the experiments) \cite{cartpole-openai}.

The episode starts with the pendulum upright, and it ends when the pendulum falls over to one of the sides. At every step, the agent receives a reward of 1 if the pendulum has not fallen to the side. The maximum number of steps is 200, which in turn limits the maximum reward obtainable to 200. A task is completed when the agent achieves a stable optimal policy. In our experiments, that translates to obtaining a reward of 199 or greater during 10 or more consecutive episodes. Evaluation is not done separately - the same trials used for training are used for evaluation, so the agent has to carefully consider the exploration-exploitation trade-off in order to achieve the goal. 


The state perceived by the agent is defined by four continuous variables: 
\begin{itemize}
    \item the cart position, ranging from -2.4 to 2.4;
    \item the cart velocity, ranging from -Inf to Inf;
    \item the pole angle, ranging from -41.8 to 41.8 degrees;
    \item the pole velocity at tip, ranging from -Inf to Inf.
\end{itemize}

The agent's performance is measured in terms of episodes to completion (ETC), defined as the number of episodes required for the agent to complete the task. The maximum ETC allowed is 1000. If the agent is unable to reach the goal within the delimited number of episodes, ETC is set to 1000 and the trial is registered as a failure. In our settings, no initial knowledge of the environment is allowed, including sampling random transitions from the environment to pre-fill a replay buffer as seen in \cite{mnih2015human}. As a consequence, the agent only starts to learn after its buffer has a number of experiences equal to or greater than the learning batch size defined.

\begin{table*}[t!]
\caption{Hyperparameters selected for the experiments.} 
\centering
\begin{threeparttable}
\begin{tabular}{||l c l||} 
 \hline
 Hyperparameter & Value & Explanation \\ [0.5ex] 
 \hline\hline
    Learning Rate ($\alpha$) & 0.001 & Step-size update for the neural network weights \\ 
    Discount Rate ($\gamma$) & 0.99 & Used to discount future rewards \\
    Soft Update Rate ($\tau$) & 0.005 & Step-size update for the target network \\
    Experience Buffer Size ($\kappa$) & 20000 & Maximum number of experiences in the experience buffer \\
    Replay Batch Size & 32 & Number of experiences sampled for each learning step \\
    Exploration Rate ($\epsilon$) Initial Value & 1.0 & Initial value for $\epsilon$-greedy exploration \\
    Exploration Rate ($\epsilon$) Final Value & 0 & Final value for $\epsilon$-greedy exploration \\
    Exploration Rate ($\epsilon$) Decay & 4000 & Number of frames over which the initial value of $\epsilon$ is linearly annealed to its final value \\ 
    Experience Transfer Batch Size ($\alpha$)\tnote{a} $^{\mathrm{a}}$& 128 & Maximum number of experiences shared at each transfer round\\ 
    Priority Replay $\alpha$ $^{\mathrm{b}}$& 0.6 & Prioritization exponent, determines how much prioritization is used  \\
    Priority Replay $\beta$ Initial Value $^{\mathrm{b}}$& 0.4 & Initial value for the importance sampling correction exponent \\
    Priority Replay $\beta$ Final Value $^{\mathrm{b}}$& 0 & Final value for the importance sampling correction exponent \\
    Priority Replay $\beta$ Decay $^{\mathrm{b}}$& 10000 & Number of frames over which the initial value of $\beta$ is linearly annealed to its final value  \\ 
    Focused ES Threshold ($\zeta$) $^{\mathrm{c}}$& 10 & Number of experiences below which the agent considers the region unexplored
     \\ [1ex]
\hline
\end{tabular}
\begin{tablenotes}
\item[a] Applied only to multiagent variants.
\item[b] Applied only to methods with priority replay.
\item[c] Applied only to methods using Focused ES.
\end{tablenotes}
\end{threeparttable}
\label{hyperparameters}
\end{table*}

\subsection{Baseline Algorithm}

To test our proposal, we conducted experiments with 6 different versions of the algorithm. Two are single agent implementations, to be used as a baseline, and four are multiagent implementations enhanced by the proposed ES methods.  

The baseline algorithm is the Deep Q-Network (DQN), which successfully demonstrated super-human performance in the Arcade Learning environment \cite{mnih2015human}. Several improvements to DQN have been introduced over the last few years, and the most important of them were considered in the baseline implementation, approaching the state-of-the-art technique. 

The most relevant modification is using the target network to accrue the value of the next state and action when bootstrapping, introduced in \cite{van2016deep}. This modification to the original DQN is introduced as a new algorithm, Double-DQN, which we will call here DQN for simplification. We've also applied soft updates to the target network, using a parameter $\tau$ which controls how much of the learning network is merged with the target network at every step \cite{lillicrap2015continuous}.

A batch of experiences is randomly sampled from the experience buffer at every step and used to calculate the loss and update the weights of the network accordingly. Exploration is done using $\epsilon$-greedy policies, with epsilon reduced at every step by a linear rate. The linear rate is calculated by setting the final epsilon value, a minimum rate of exploration, and a number of frames to decay. The epsilon decay rate is given by the number of frames divided by the initial epsilon minus the final epsilon. 

To approximate the action-value function we implement a Multilayer Perceptron, with one input layer, two hidden layers, and an output layer. As in DQN, the neural network approximates the action-value function $Q$, mapping a state to action values. The input layer has four neurons, equivalent to the state size, and the output layer has two neurons, equivalent to the number of actions. There are two hidden layers of 16 and 8 neurons respectively, which uses rectified linear units as the non-linear activation function. This architecture is a modification of the original DQN publication \cite{mnih2015human}, with significantly fewer degrees of freedom due to the simplicity of the task. No other variants of neural networks architectures were tested, as it is not the focus of this work. 

Two versions of the baseline are used. The first is the DQN, as described above. The second, which we call DQN-PR, uses Prioritized Replay to decide which samples to replay at every learning step. In DQN-PR, each sample is assigned a maximum priority when entering the batch, ensuring it is sampled at least once. Every time an experience is sampled, its priority is updated according to the TD-error calculated for it. The TD-error represents how much of an impact an experience had in the weight adjustment done in a particular step. It is also a proxy for how surprised the agent is in experiencing that transition. Its implementation is inspired by neuroscience studies reporting similar behavior in rodents. \cite{schaul2015prioritized} 

The neural network implements the Adam optimizer, and its parameters, $\beta_1$ and $\beta_2$, are set to the default values $\beta_1 = 0.9$ and $\beta_2 = 0.999$ considered optimal for the majority of problems regarding neural networks \cite{kingma2014adam}. Clipping the gradients for the neural network was also attempted, but it led to inferior performance and was not considered in the implemented baseline.  The remaining hyperparameters were optimized by grid search. The complete list of hyperparameters selected for the baseline algorithm is given in Table \ref{hyperparameters}.

\subsection{Experimental Procedures}

Each repetition is called a trial. A trial ends when all agents complete the task. Directly comparing two algorithms in a single trial is not reliable due to the non-deterministic nature of both the agent's function and the environment function. The agent's action selection, experience buffer sampling, and neural network initialization are all in part stochastic processes. The environment's transition function is given by a probability distribution. Therefore, in order to compare two or more algorithms, the experiment is repeated a number of times with different random seeds, and the distribution of the results are compared, as proposed in \cite{henderson2017deep}.

The main performance metric used for evaluation is ETC. In the single agent variant (baseline), a trial adds only one sample to the distribution. In the multiagent variant, the results of all agents are added to the distribution. The number of trials executed is 100 for single agent and 50 for multiagent variant with two agents; as a consequence, the distribution for each variant tested is composed of 100 samples. 

The size of the distribution is enough to be considered representative of the entire population. Although 30 is typically considered to be the minimum sample size required to apply large-sample statistics, considering the high variance of the sample results and seeking to provide robust outcomes, we've decided on using 100 as the sample size.  

\subsection{Results and Discussion}

With the experimentation procedure explained, we proceed to present and discuss the results achieved \footnote{Source code for the experiments and detailed results are available at https://github.com/lucasosouza/fasterRL}. 

We first compare multiagent variants DQN + Naive ES and DQN + Focused ES with single agent DQN baseline, and multiagent variants DQN-PR + Prioritized ES and DQN-PR + Prioritized Focused ES with single agent DQN-PR baseline. We aim to show that the cooperative multiagent variants can outperform the single agent baseline. 

\begin{figure}[!ht]
\centering
  \includegraphics[width=1\linewidth]{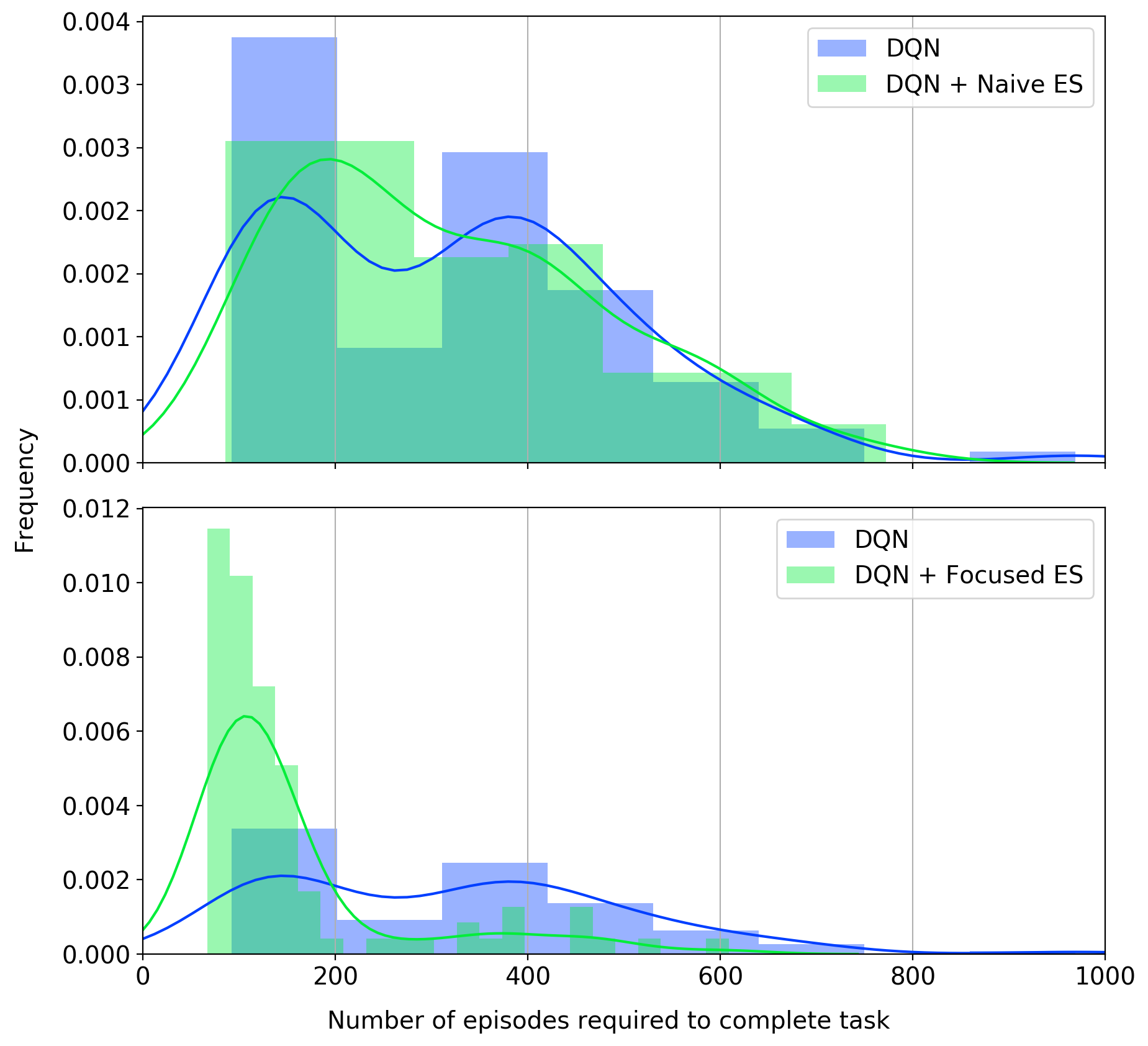}\\
  \caption{Comparison of single agent DQN with multiagent DQN with naive and Focused ES.}
  \label{fig:singleVSmulti}
\end{figure}

In Fig. \ref{fig:singleVSmulti} we plot the samples from a single agent DQN versus a multiagent DQN with two agents sharing experiences. In all experiments, a normalized histogram and a kernel density estimation of both distributions are used to compare. 

Results show that multiagent DQN with Naive ES adds no improvement over the single agent DQN. However, multiagent DQN with Focused ES shows a significant improvement over the baseline. The Focused ES method has an average of 154 and a standard deviation of 112 ETC, compared to an average of 318 and a standard deviation of 177 ETC in single agent DQN, resulting in a 51.4\% improvement in performance. A two sample K-S test rejects the null hypothesis that both samples are drawn from the same distribution, with a p-value of $3.70 \times 10^{-12}$.

\begin{figure}[!h]
\centering
  \includegraphics[width=1\linewidth]{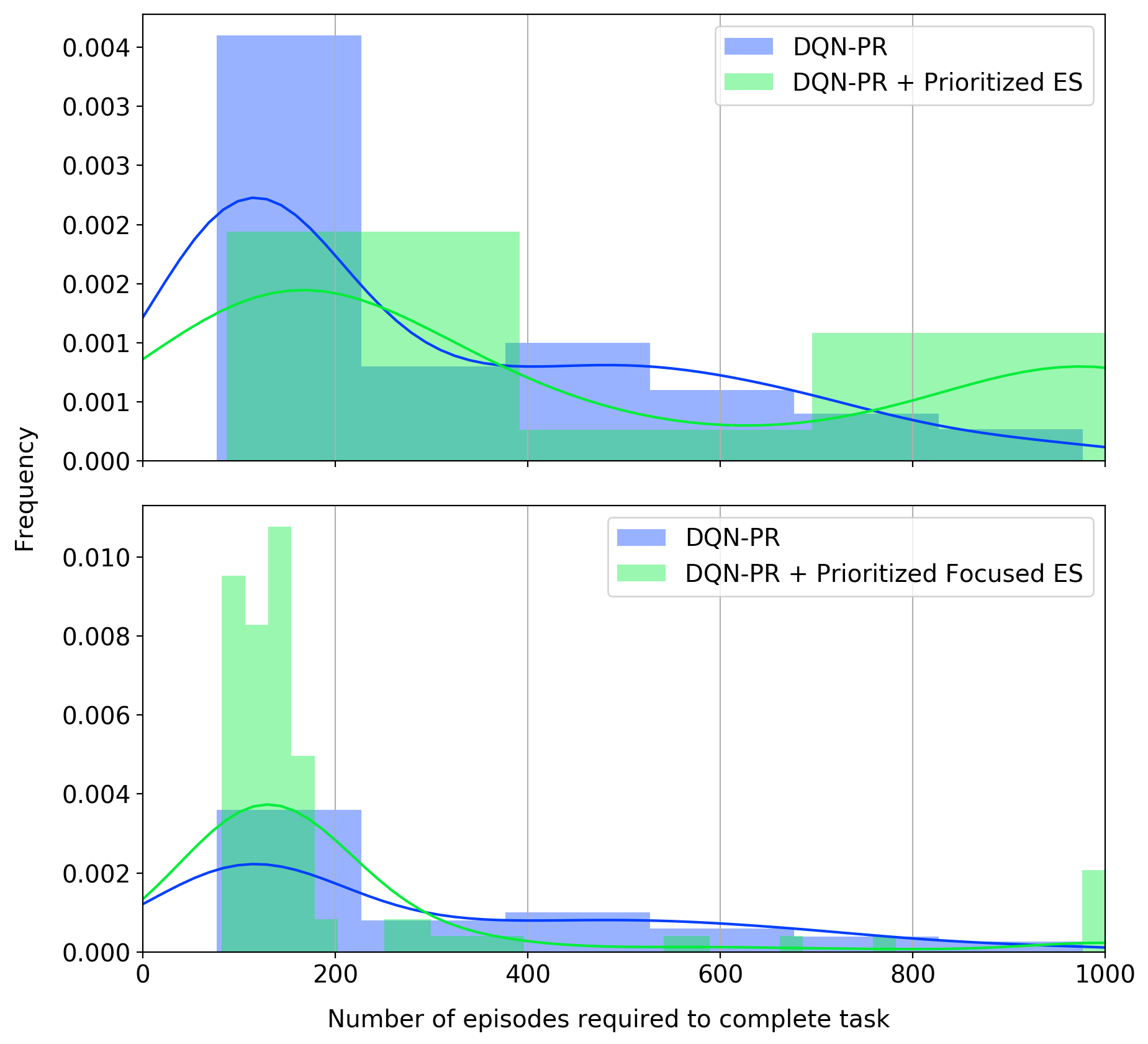}\\
  \caption{Comparison of single agent DQN-PR with multiagent DQN-PR with prioritized and Prioritized Focused ES.}
  \label{fig:singleVSmulti-prio}
\end{figure}

The same comparison is shown for the DQN-PR algorithms in Fig. \ref{fig:singleVSmulti-prio}. Multiagent DQN-PR with Prioritized ES has a significantly lower performance compared to single agent DQN-PR, with 26 out of 100 samples failing to complete the task. We speculate that the regular stream of new experiences with maximum priority prevents the agent from replaying old experiences, which are eventually discarded before they can be reused for learning when the buffer reaches maximum capacity. 

By combining the Focused ES method with Prioritized ES we can ensure only the most relevant experience are shared. DQN-PR with Prioritized Focused ES shows an improvement in performance of 31.1\% over the baseline DQN-PR. As before, we apply a K-S test to test the hypothesis of both samples being drawn from the same distribution, which we reject with a p-value of $1.74 \times 10^{-4}$. 

\begin{figure}[!t]
\centering
  \includegraphics[width=1\linewidth]{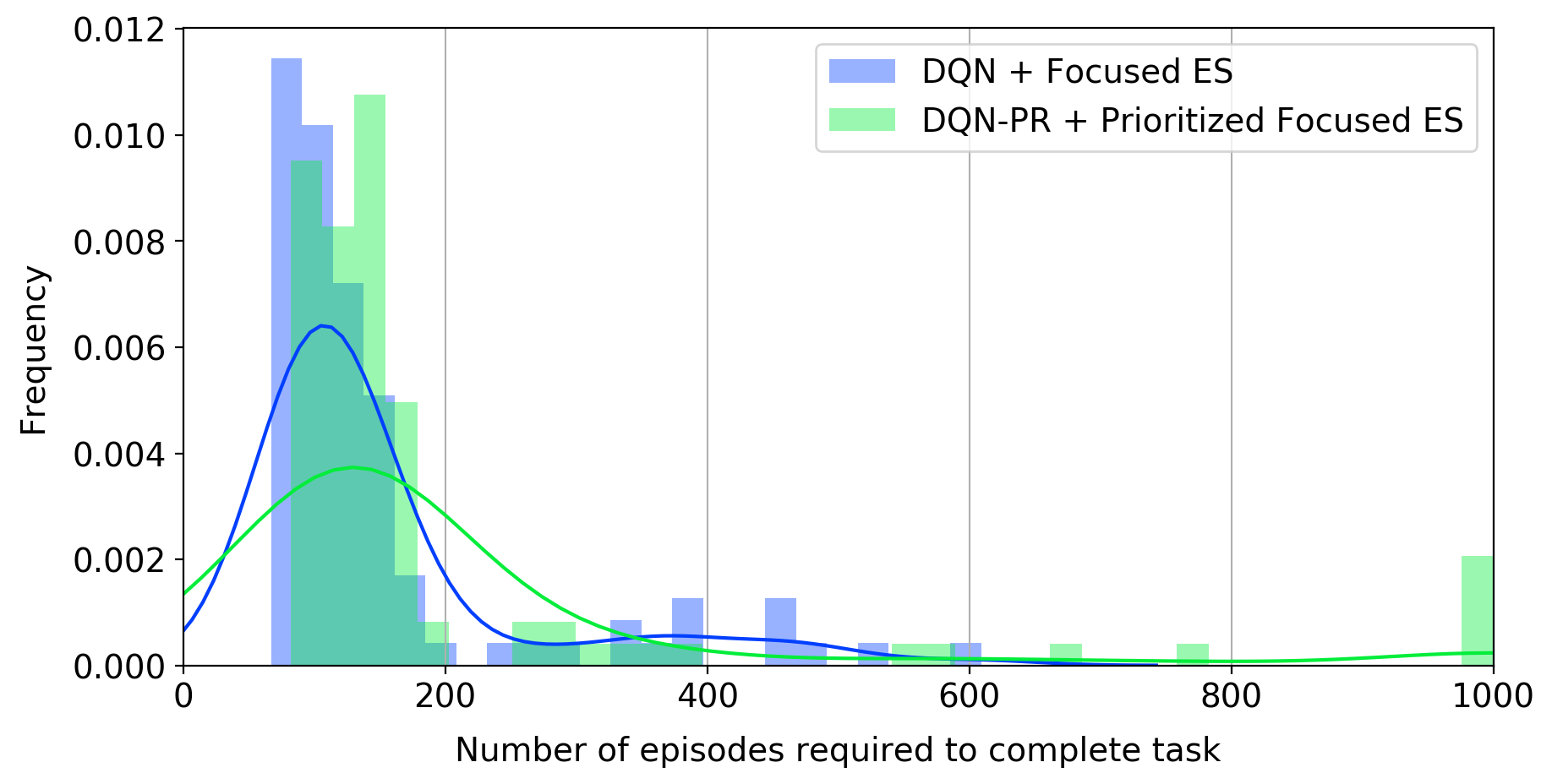}\\
  \caption{Comparison of multiagent variants DQN and DQN-PR with Focused ES.}
  \label{fig:multiComparison}
\end{figure}

We directly compare the best approaches DQN + Focused ES with DQN-PR + Prioritized Focused ES in Fig. \ref{fig:multiComparison}. DQN + Focused ES distribution shows lower average and lower variance, while DQN-PR + Prioritized Focused shows a long right-side tail distribution which pulls the average higher, with the agent failing to achieve the goal in 5 out of 100 samples. The discussed measures and other statistics for all algorithms tested are presented in Table \ref{table:results}.

\begin{table}[htbp]
\caption{ETC and number of failed trials per method}
\begin{center}
\begin{tabular}{|l|c|c|c|c|}
    \hline
    Method &  
    \multicolumn{1}{|p{0.95cm}|}{\centering ETC \\ Mean} &
    \multicolumn{1}{|p{0.95cm}|}{\centering ETC \\ Deviation} &
    \multicolumn{1}{|p{0.7cm}|}{\centering Trials \\ Failed} &
    \multicolumn{1}{|p{1.33cm}|}{\centering ETC \\ Improvement} \\
    \hline\hline
    DQN                             &  317.65   &  176.64        &   0            &   -           \\
    \hline
    DQN + Naive ES                  &  318.19   &  163.26        &   0            &  -0.2\%           \\
    \hline
    DQN + Foc. ES                &  154.44   &  111.92        &   0            &  +51.4\%           \\
    \hline
    DQN-PR                          &  300.22   &  246.05        &   0            &   -           \\
    \hline
    DQN-PR + Pr. ES         &  459.63   &  370.57        &  26            & -53.1\%           \\
    \hline    
    DQN-PR + Pr.Foc. ES &  206.87   &  214.63        &   5            &  +31.1\%           \\
    \hline
\end{tabular}
\label{table:results}
\end{center}
\end{table}

\begin{figure}[t]
\centering
  \includegraphics[width=1\linewidth]{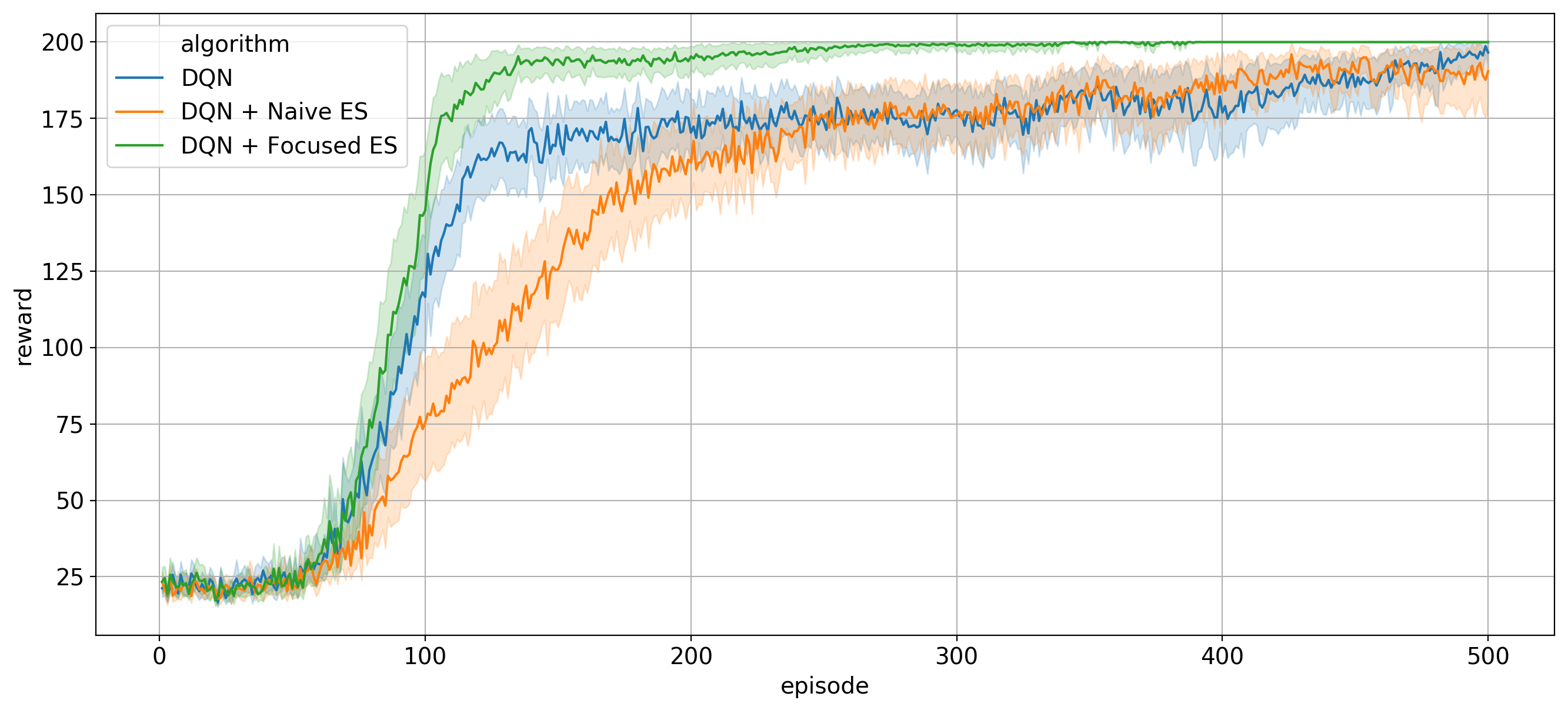}\\
  \caption{Episode reward evolution in DQN.}
  \label{fig:rewards-regulardqn}
\end{figure}

We can better understand how Focused ES affects learning by analyzing the learning dynamics episode by episode. Figs. 6 and 7 plot the average reward along with episodes for the considered algorithms. In the plots, each line represents an average over 100 trials (with standard deviation shown as shaded lines). To enhance presentation, all lines are shown up to 500 episodes\footnote{In the case of trials that completed the task before 500 episodes, we consider the last obtained reward to compute the average of subsequent episodes.}. In Fig. \ref{fig:rewards-regulardqn} we see how in DQN + Focused ES learning progress faster right from the beginning, while in DQN + Naive ES progress is slower, even when compared to the single-agent variant. The differences in performance are most notable after they reached a high level reward, around 175. Most of the variance in ETC in DQN and DQN + Naive ES can be explained by the time it takes to cover the last steps towards the target, leaping to the maximum reward of 200. DQN + Focused ES is able to overcome this last stage using fewer episodes. 

\begin{figure}[t]
\centering
  \includegraphics[width=1\linewidth]{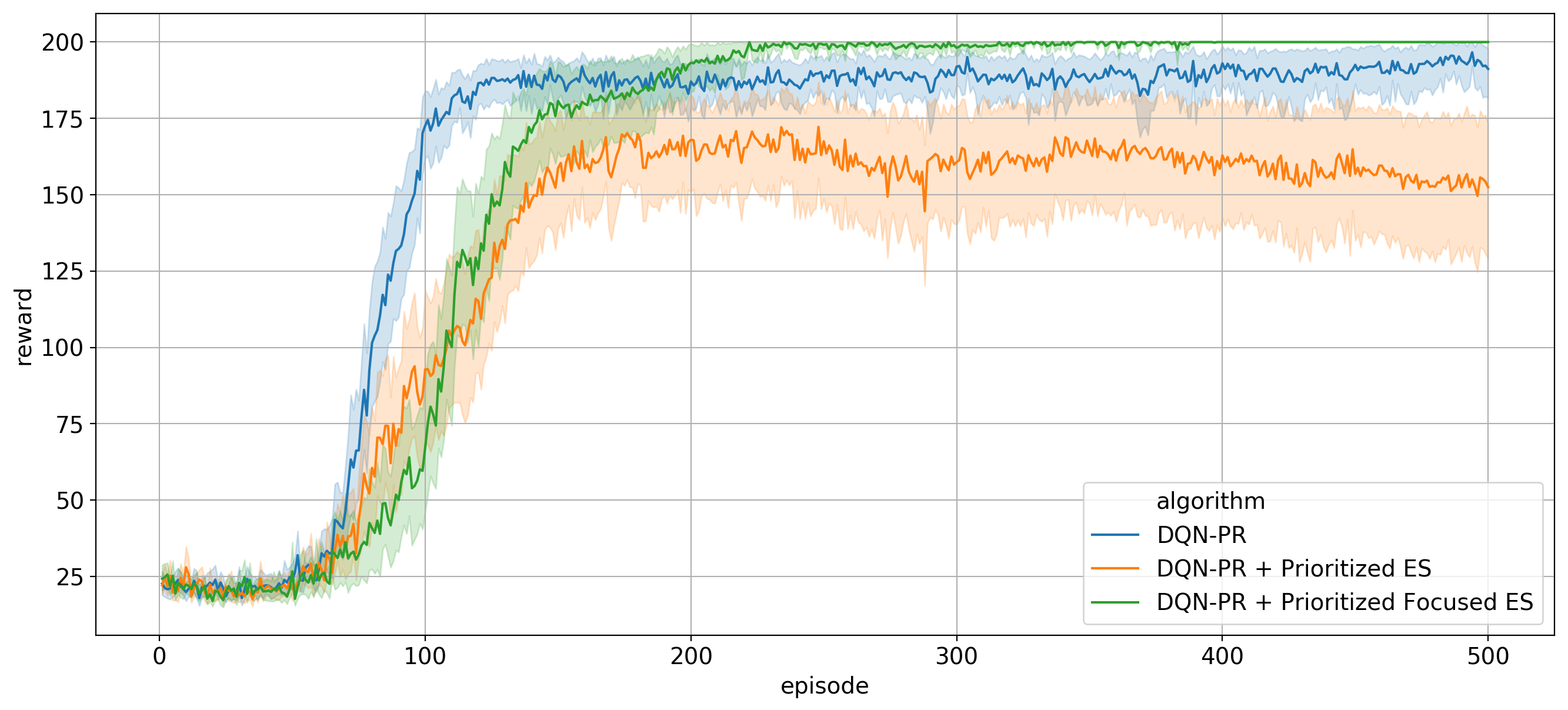}\\
  \caption{Episode reward evolution in DQN-PR.}
  \label{fig:rewards-priodqn}
\end{figure}

Similar behavior occurs when priority replay is added, seen in Fig. \ref{fig:rewards-priodqn}. In this case, DQN-PR + Prioritized Focused ES has slower learning compared to the single agent variant in the first 180 episodes. However, as with regular DQN, the single agent variant plateaus in the last step of progression, while the Focused ES version is able to continue progressing towards the optimal policy. 


\begin{figure}[t]
\centering
  \includegraphics[width=1\linewidth]{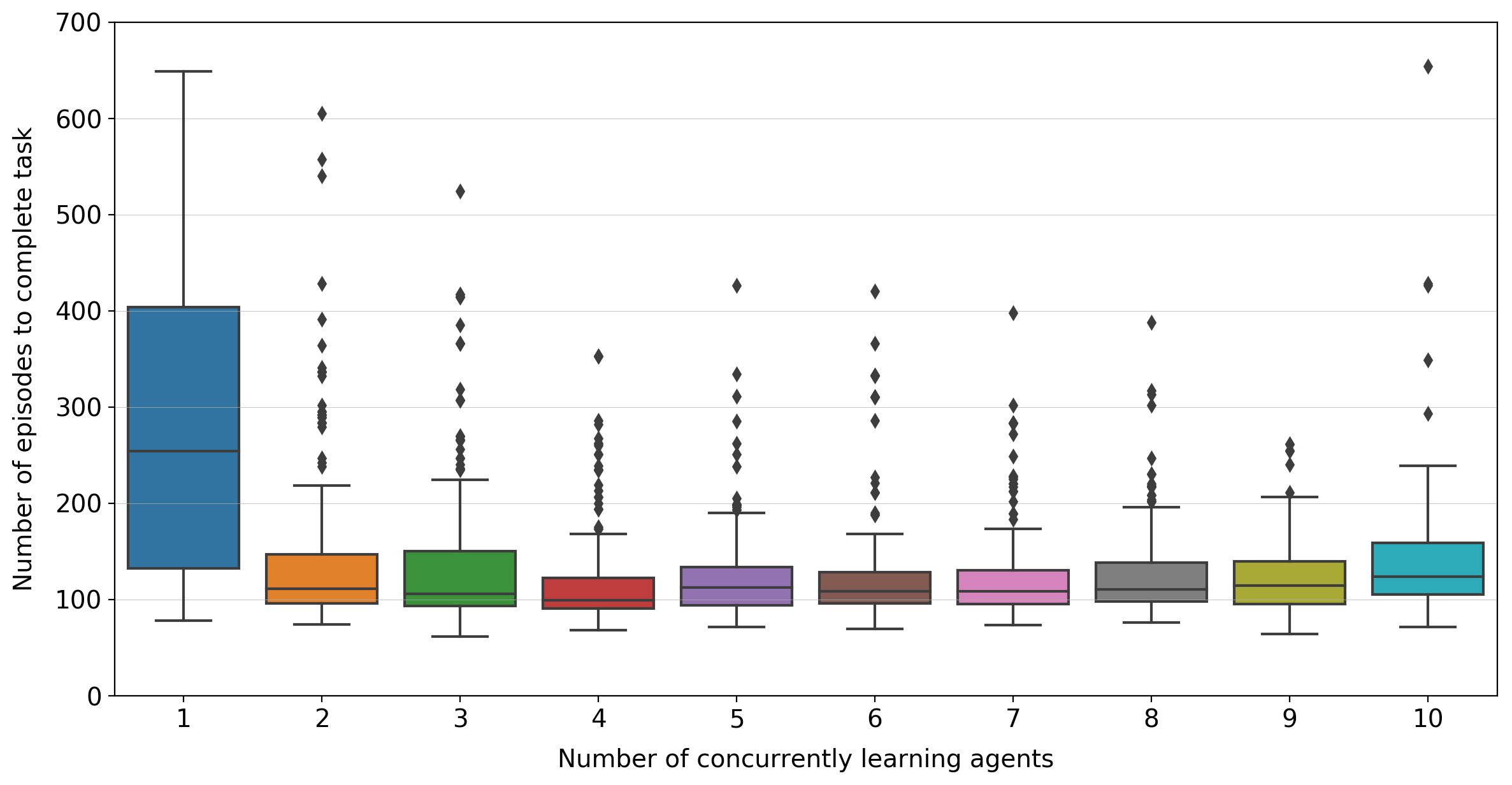} \\
  \caption{Boxplots showing the distribution of the number of episodes to complete task with different numbers of concurrently learning agents.}
  \label{fig:multiagentquantiles}
\end{figure}

Furthermore, we test if adding more agents to the multiagent variant using DQN + Focused ES can increase the performance. The results are shown in Fig. \ref{fig:multiagentquantiles} and Table \ref{table:multiagentquantiles}. The biggest impact occurs when adding a second agent to the experiment. There is a small improvement up to four agents. After five agents, performance starts to decrease, with ten agents experiment showing inferior performance compared to the two agents. Each transfer is limited by a number of experiences $\kappa$, but we did not set an upper limit to the total number of experiences received in an episode when transfer from all agents are considered. Adding more agents increase the total number of experiences received by an agent, which after a point stops being helpful. Adding diverse off-policy experiences to the buffer increases exploration, and we speculate there is an upper bound of how much exploration can be increased through this approach before it impacts the learning process.

\begin{table}[t]
\caption{Q1, Q2 and Q3 ETC for Focused Experience Sharing between 1 to 10 agents}
\begin{center}
\setlength\tabcolsep{4.5pt}
\begin{tabular}{|l|c|c|c|c|c|c|c|c|c|c|}
\hline
Quantile &  1   &  2   &  3   &  4   &  5   &  6   &  7   &  8   &  9   &  10  \\
\hline\hline
Q(.25) &  132 &   96 &   93 &   90 &   94 &   96 &   95 &   98 &   95 &  105 \\
\hline
Q(.5) &  254 &  111 &  105 &   99 &  112 &  108 &  108 &  110 &  114 &  123 \\
\hline
Q(.75) &  403 &  146 &  149 &  122 &  133 &  128 &  130 &  138 &  139 &  158 \\
\hline
\end{tabular}
\label{table:multiagentquantiles}
\end{center}
\end{table}




\section{Related Work}

Previous work in experience sharing has been explored in discrete state space problems \cite{tan1993multi, nguyen2018swarm}. In those, tabular representations of the Q-value function allowed the action-value to be directly transferred between tables, allowing for faster learning. In continuous or complex state spaces, the $Q$ function can no longer be represented by a table, so function approximators are used instead. Approximating the action-value function rules out the possibility of directly transferring action values between two agents.

The student-teacher approach to the RL multiagent setting is explored in \cite{torrey2013teaching}. It is extended in \cite{da2017simultaneously} which proposes a framework where concurrently learning agents can either play the role of learner or teacher, similar to the approach we adopted. The knowledge is shared in the form of action advice from a teacher which overrides the action selection of the student.

Although action advice has proven to increase performance in cooperative multiagent settings, it requires instantaneous communication between agents, with atomic information of one transition shared in each communication. In our proposal, communication is only done once at the end of each episode, and knowledge regarding several transitions can be packed into a single communication effort, making it more realistically applicable to real-world problems. Another main difference is we focus on what to share, defining a method to select the experiences which can be most useful for the student learning, while \cite{torrey2013teaching, da2017simultaneously} focus on when to share.

Also related is the research conducted in distributed learning using DQN \cite{ong2015distributed, nair2015massively}. They introduce fully distributed learning algorithms, leveraging the DistBelief software framework \cite{dean2012large} to train a neural network in a distributed approach. In \cite{horgan2018distributed} the agents have independent action-value networks, but a centralized ER buffer, highlighting the impact of buffer diversity in improving learning.  These investigations use a hybrid approach with partially centralized learning systems, while we focus on fully independent agents who only communicate on an episode by episode basis.


\section{Conclusion}
\label{sec:conclusions}

In this paper, we proposed and evaluated four methods to accelerate learning in cooperative, multiagent deep reinforcement learning settings. Through an empirical analysis, we demonstrate that experience sharing between two concurrently learning agents does not improve the agents' performance. We then propose a novel method, called Focused ES, that decreases the number of episodes required to complete a task by a factor of two.

Our method can be readily deployed in applications using concurrently learning RL agents, halving the learning time of an agent just by reusing experiences learned by its peers. As opposed to previous approaches, our method does not require a centralized neural network or a centralized buffer, making it more easily extendable to industrial applications where the latency and bandwidth of communication between agents are limited. 

In spite of the promising results, we envision interesting directions for future work. 
One limitation of our work is its applicability to homogeneous agents and non-dynamic environments. Recent results on ES techniques to dynamic environments and heterogeneous agents \cite{verstraetenreinforcement, garant2017context} can be combined with our Focused ES algorithm. A further improvement is to extend the learning setting to Markov games, in which an agent's interaction makes the environment non-stationary, a typical formulation of multiagent RL problems. Those limitations can be addressed in future work. 









\section*{Acknowledgment}

We thank the anonymous reviewers for their valuable suggestions. 
Gabriel was partially supported by FAPERGS (grant number 19/2551-0001277-2). 
Celia was partially supported by CNPq through a PQ-2 research productivity grant (number 303863/2015-3).

\bibliographystyle{IEEEtran}
\bibliography{IEEEabrv,references}

\end{document}